\documentclass{article}

\usepackage{courier} 

\usepackage{natbib}

\setcitestyle{authoryear,round,citesep={;},aysep={},yysep={,},notesep={, }}

\usepackage[utf8x]{inputenc}
\usepackage[T1]{fontenc}

\usepackage{graphicx}

\usepackage{subfigure}   

\usepackage[unicode=true]{hyperref}
\hypersetup{colorlinks=true,breaklinks,    
            linktoc=all,
            urlcolor=[rgb]{0.23,0.55,0.7},
            linkcolor=[rgb]{0.4,0.4,0.4},
            linkbordercolor=[rgb]{1.0,0.0,0.0},
            citecolor=[rgb]{0.4,0.4,0.4},
            pdfborderstyle={/S/U/W 0}}

\usepackage{amsmath}
\usepackage{url}

\DeclareMathOperator{\sign}{sgn}

\date{\today}
\title{Confusing Deep Convolution Networks by Relabelling}
\author{
  Leigh Robinson\\
  \texttt{leigh.robinson@warwick.ac.uk}
  \hfill \break
  \and
  Ben Graham\\
  \texttt{b.graham@warwick.ac.uk}
}

\begin{document}
\maketitle

\begin{abstract}
\noindent Deep convolutional neural networks have become the gold standard for image recognition tasks, demonstrating many current state-of-the-art results and even achieving near-human level performance on some tasks. Despite this fact it has been shown that their strong generalisation qualities can be fooled to misclassify previously correctly classified natural images and give erroneous high confidence classifications to nonsense synthetic images. In this paper we extend this work and present a straightforward way to perturb any image in such a way as to cause it to acquire any other label from within the dataset while leaving this perturbed image visually indistinguishable from the original.
\end{abstract}

\section{Introduction\label{sec:Introduction}}
Deep convolutional neural networks are currently at the forefront of image classification tasks, reporting many state-of-the-art results, and indeed near-human performance on large scale natural image benchmarks in recent years \citep{Krizhevsky:2012,Szegedy:2014GD,Schroff2015,He2015}. One longstanding assumption was that since these networks exhibit strong generalisation qualities to unseen test images they should also be stable with regards to small perturbations of the input image - since such an image undoubtably still contains an exemplar of the same class. This notion of stability has in fact been shown to be false.

Work by \citet{Szegedy:2014wr} was first to show that it is indeed possible to find small perturbations of a given image (via a L-BFGS optimisation process) that leave them visually indistinguishable from the original, but nonetheless cause the network to misclassify the previously correctly classified exemplar. Moreover these adversarial exemplars (the perturbed images) were shown to generalise well across different classifiers that had different architectures. A similar result for generated images has been shown by \citet{Nguyen:2014wn}. In this instance instead of starting with an exemplar, the optimisation proceeds either from a random noise image or simply directly generates a synthetic image with geometric patterns that are strongly classified as a particular class. 

A plausible explaination for the ease at which these networks can be fooled is given by \citet{goodfellow2014-adversarial}. With the central problem being identified as the move towards using somewhat linear functions at each layer due to the ease at which they can be trained. This results in the representation space being partitioned into halfspaces of increasingly confident but erroneous class predictions as you move away from the distribution of the training exemplars.

In this paper an extension of previous work by \citet{goodfellow2014-adversarial} is presented that shows that small perturbations of a correctly classified image can be constructed in an efficient way that allows for the original image to be re-assigned to any of the other classes while remaining visually indistinguisible. Moreover this process is robust with respect to both initial and target class. In Section \ref{sec:Implementation} the implementation details of the relabelling process is discussed. In Section \ref{sec:image_relabelling} the results of relabelling experiments done on the ImageNet dataset \citep{imagenet_cvpr09} are presented and some of the discovered limitations of the relabelling process are discussed. Section \ref{sec:feature_analysis} adds some analysis of how the relabelling manifests as changes to the representations of the intermediate layers of the network.
\section{Implementation\label{sec:Implementation}}
To explore the possibility that an input image (that on a trained network is correctly classified) can be re-assigned to another label experiments were performed with the Caffe framework \citep{jia2014caffe}\footnote{available at: \url{https://github.com/BVLC/caffe}}. The Caffe framework is particularly convenient for two reasons: The framework has publically available pre-trained implementations of "AlexNet" \citep{Krizhevsky:2012} and "GoogLeNet" \citep{Szegedy:2014GD} and straightforward modifications of the model definitions allow for the backpropagation when computing the gradients to proceed right up to the input image layer. These gradients are then used to update the input image in a batch of gradient descent updates.

Formally, given a network model with parameters $\theta$, an input image $X$, a "confusing" target label $y$ and a cost function, $L(\theta,X,y)$. In the networks considered in this paper, $L$ is the standard softmax function. When computing the set of gradients, $\frac{\partial\mathrm{L}}{\partial\mathrm{\theta}}$, we backpropagate one step further to that of the input image X, yielding $\frac{\partial\mathrm{L}}{\partial\mathrm{X}}$. The process then proceeds like traditional gradient descent, with $N$ gradient updates occuring to the input image, $X$ as follows

\begin{equation}
X_{i+1} \leftarrow  X_{i} - \alpha f(\frac{\partial\mathrm{L}}{\partial\mathrm{X_{i}}})
\end{equation}

\begin{figure}[ht]
    \centering
    \subfigure[$\frac{\partial\mathrm{L}}{\partial\mathrm{X}}$\label{fig:gradient_raw}]{\includegraphics[width = 0.3 \textwidth]{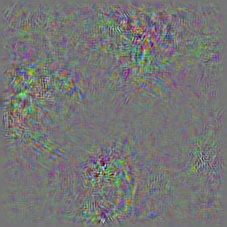}}\hspace{4em}
    \subfigure[$\sign (\frac{\partial\mathrm{L}}{\partial\mathrm{X}}) $\label{fig:gradient_sign}]{\includegraphics[width=0.3\textwidth]{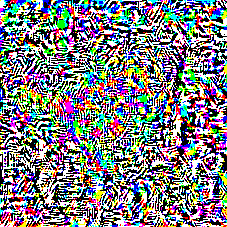}}
    \caption{This figure shows the appearance of the gradients that are computed on the first iteration when relabelling an image from class 10, to class 348. Notice that the for the gradient passed through the $\sign$ function (\ref{fig:gradient_sign}) the values are much more spread out spatially, and each colour channel are bounded between [-1/255,1/255]. This stops "hot spots" appearing in the image from the large clustered values apparent in \ref{fig:gradient_raw} when used in the gradient descent.\label{fig:gradients}}
\end{figure}

The only remaining definition is the nature of the function, $f$ in the above equation. Initial testing showed that while using the raw gradient allowed for a succesful relabelling of the input image, it was a destructive process leaving the image with noticable blemishes unless very careful tuning of the update step, $\alpha$ and the number of total iterations was undertaken. The problem with the raw gradient (as seen in Figure \ref{fig:gradients}) is that there are strong "hot-spots" where the gradient values have a much larger magnitude than the surrounding average value. This causes local regions of the image to be greatly modified at each update step. As mentioned above this problem can be somewhat ameliorated by scaling the gradient very small (i.e small $\alpha$) while greatly increasing the number of updates. However this is not ideal as it not only results in much longer computation times but also makes the process quite brittle - as a key constraint is that the resulting relabelled images remain visually indistinguishable from the originals. Following the example of \citet{goodfellow2014-adversarial} rescaling the gradient so that the relative strengths of the differing spatial locations is much less severe is beneficial. This was accomplished by setting the function $f$ to be the signum which is then scaled by the quantisation factor of standard 8-bit images, specifically
\begin{equation}
f(X) = \sign (X) / 255.0
\end{equation}
\section{Image relabelling\label{sec:image_relabelling}}
To perform image relabelling we simply start the optimisation from the original image, and provide a target label that we wish to assign the original image.

Simlarly to \citet{Szegedy:2014wr} we use the distortion measure defined as 
\begin{equation}
distortion(x^{'}, x) = \sqrt{\frac{\sum_n (x_i^{'} - x_i)^2}{n}}
\end{equation}
where $x$ is the original image, $x^{'}$ the relabelled image and $n = 50176$ is the number of pixels in the image.

\begin{figure}[!ht]
    \centering
    \includegraphics[width=0.9\columnwidth]{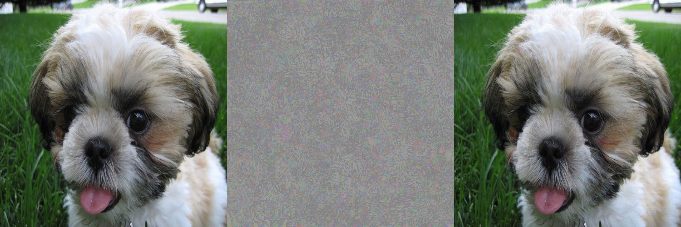}
    \par\vspace{0.5 mm}
    \includegraphics[width=0.9\columnwidth]{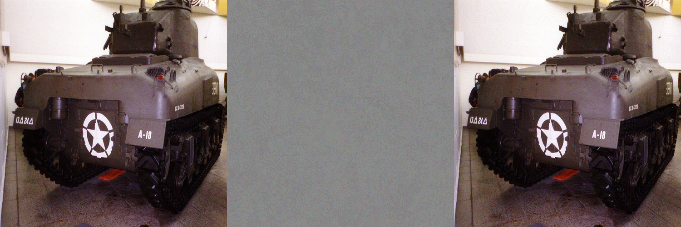}
    \caption{Examples of class relabelling on GoogLeNet. Left: Original images that are correctly classified as `Shih-Tzu' and `Half-track'. Right: relabelled images such that the classification output is reversed - i.e the `Shih-Tzu' is now strongly classified as `Half-track' and vice-versa. Centre: Pixel differences multiplied by 10 and scaled to mean-level for visibility. The distortion introduced in the top relableling was 0.00864 and the bottom 0.00712. These are a random pair of images taken from the above classes that were chosen to be qualitatively "maximally different". As we show in Section \ref{sec:subsec:how_robust} relabelling is not sensitive to the initial or target class.\label{fig:relabelling_examples}}
\end{figure}

\begin{figure}[!ht]
    \centering
    \subfigure[Half track exemplar before relabelling.\label{fig:relabelling_probs:a}]{\includegraphics{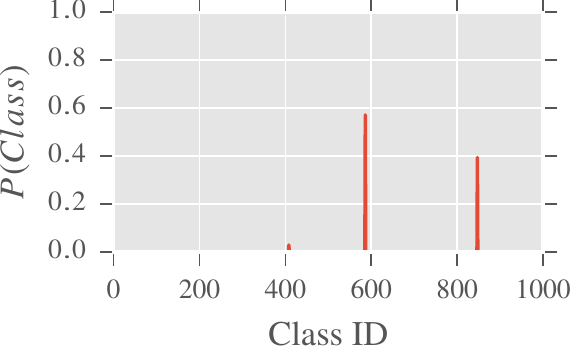}}
    \subfigure[Half track exemplar after relabelling.\label{fig:relabelling_probs:b}]{\includegraphics{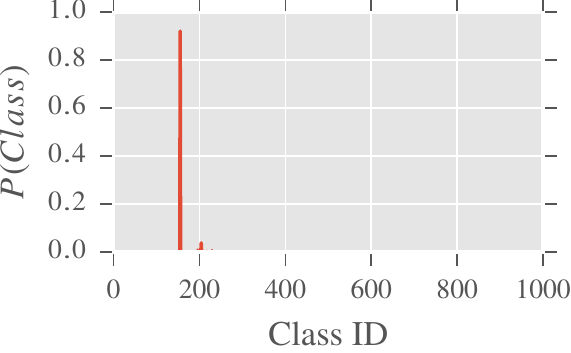}}
    \subfigure[ShihTzu exemplar before relabelling.\label{fig:relabelling_probs:c}]{\includegraphics{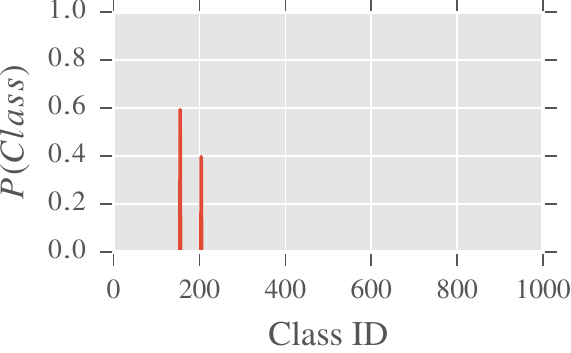}}
    \subfigure[ShihTzu exemplar after relabelling.\label{fig:relabelling_probs:d}]{\includegraphics{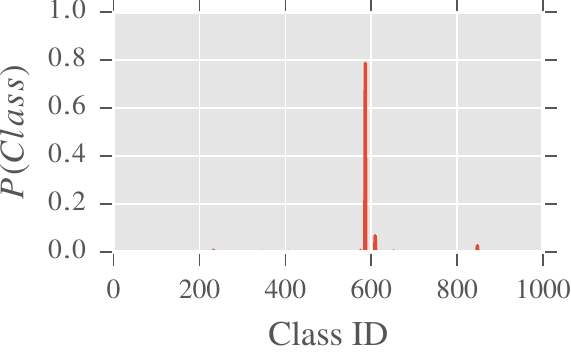}}
    \caption{The class probabilities for each of the four images in Fig \ref{fig:relabelling_examples}. Notice that after relabelling the uncertainty is also reduced. \label{fig:relabelling_probs}}
\end{figure}

In Figure \ref{fig:relabelling_examples} we show the results of the above image relabelling process on GoogLeNet, by choosing two exemplars at random from the ImageNet classes `Shih-Tzu' and `Half-Track' and swapping their labels. Average distortion measures from 30 such relabelling procedures was found to be 0.00814, which leaves the altered images visually indistingusible from the originals. Figure \ref{fig:relabelling_probs} shows the associated class probabilities, confirming that the relabelling process has worked successfully. Furthermore the uncertainty of the relabelled image label is consistently very low when compared to that of true images of the class.

It should be noted that the classes were chosen to ensure that the initial experiments were on images that had significant qualitative features (consider the differences in texture, dominant features, shapes, etc exhibited by exemplars of these class) - so intuitively should be a `hard' relabelling task and expressly not due to any specific pre-screening process.
\subsection{How robust is this process?\label{sec:subsec:how_robust}}
To see how applicable this process is across the entire 1000 image classes of ImageNet, the following experiment was conducted. For each of the 1000 image classes 10 random exemplars were chosen and relabelled to each of the 999 other classes, with the caveat that the maximal distortion allowed for any relabelling attempt be capped at 0.01 (see Figure \ref{fig:relabelling_examples} for examples of such distortion values). This further restriction forces the relabelling to only be considered a success if the altered image is still visually indistinguisible and is particularly stringent in this regard as images that are corrupted with random noise which results in distortion values magnitudes larger than 0.01, while obviously altered, are still easily human-recognisible as the original class.

The results of this experiment were encouraging, with 98.7\% of all class/target pairs were successful at being re-labeled below the 0.01 distortion threshold. In fact no class/target pairs were significantly harder to relabel at this distortion level. This implies that it would be perfectly possible to construct an augmented version of the ImageNet test set that would attain close to 0\% accuracy, yet look visually indistinguisible from the original.

The only parameters in the relabelling process are the update stepsize, $\alpha$ and the number of iterations, $N$ (which itself is a function of the maximal distortion you will accept since the $\sign(\frac{\partial\mathrm{L}}{\partial\mathrm{X}})$ is bounded at each step). In all the simulations described in this paper $\alpha = 50$ and the number of iterations were tuned accordingly if a target distortion was required, or the process was simply iterated until the relabelling process was sucessful. It is interesting to note that the gradient is not \textit{stationary}. Simply taking the first gradient and moving an equivalent distance in image space that would have arisen from many smaller steps does not successfully re-label the image (though can often produce a mis-classificaion). This tends to suggest that there is not a simple linear relationship between the original image and the nearby perturbation that results in the new target label being acquired.

Unfortunately there are some limitations to this relabelling scheme. The changes to the image label are not robust to transformations of the image; cropping, translating and mirroring the image results in the label reverting to the correct one, which suggests that the perturbations are leveraging specific spatial location in the process of relabelling. Furthermore, while the relabelling process itself works with both GoogLeNet and AlexNet architectures (and probably many similarly architected networks), images that are relabelled on one do not transfer over to the other. All the experiments thus described have been carried out on the GoogLeNet architecture, but similar results were obtained with the AlexNet architecture and have been omitted for brevity.

\subsection{Gaussian image synthesis\label{sec:subsec:gaussian_synthesis}}
\begin{figure}[!ht]
    \centering
    \subfigure[Random gaussian image before relabelling.\label{fig:relabelling_gaussian_exemplar_before:c}]{\includegraphics[width = 0.3 \textwidth]{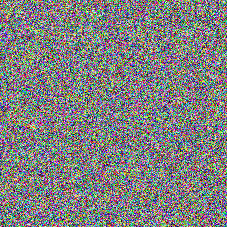}}\hspace{4em}
    \subfigure[Random gaussian image after relabelling.\label{fig:relabelling_gaussian_exemplar_after:d}]{\includegraphics[width=0.3\textwidth]{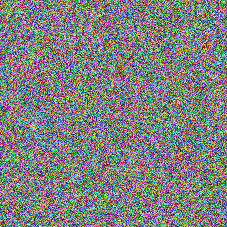}}
    \subfigure[The class probabilities of the image before relabelling.\label{fig:relabelling_gaussian_probs_before:a}]{\includegraphics{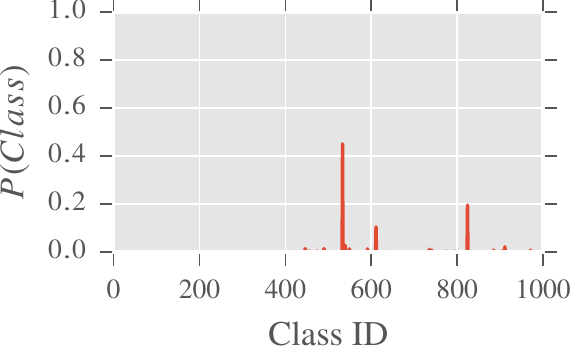}}\hspace{1em}
    \subfigure[The class probabilities after relabelling.\label{fig:relabelling__gaussian_probs_after:b}]{\includegraphics{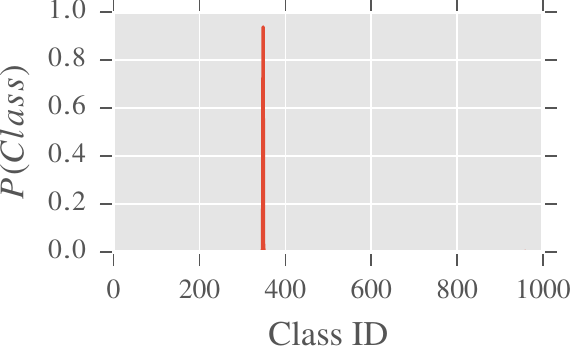}}
    \caption{These figures show a random gaussian image (\ref{fig:relabelling_gaussian_exemplar_before:c}) and the same image after being re-labeled to class 348 (\ref{fig:relabelling_gaussian_exemplar_after:d}). The bottom two graphs show the class probabilities before and after the relabelling process. The distortion introduced in this relabelling process was 0.03 - which is much higher than typically required to relabel natural images.\label{fig:gaussian_images_relabel}}
\end{figure}

This relabelling process can also generate exemplars that are completely random images devoid of human-recognisible structure, that are nonetheless classified with a high confidence by simply starting the process with a random image. Such a set of images can be shown in Figure \ref{fig:gaussian_images_relabel}.
\section{Feature analysis\label{sec:feature_analysis}}
In an attempt to understand empirically what aspect of the network is failing we have done some cursory testing of the representations of the network

A natural question to ask might be, are there any differences in the feature space representation (i.e filter activations) between images that are relabelled and those that are "true" members of the target class? To try and investigate this at differing layers within GoogLeNet a SVM (with RBF kernel) was used on the raw features generated before the final fully connected layer. Specifically, 50 images from random classes were relabelled to a single target class. Twenty of these relabelled images, alongside 20 actual exemplars from the target class were passed through GoogLeNet and the intermediary layer activations were recorded. The features generated by these 40 images were used to train an SVM to see if the impostor relabelled images could be succesfully identified from these feature representations alone. With 60 test features (30 of each type) the SVN was succesfully able to identify the relabelled images with an average accuracy of 89.6\% over 10 random sets of images and targets.

This is interesting because it supports the idea that the relabelled images are illiciting semantic "super responses" in the higher layer features. For example, if we were to relabel an image of a dog to that of a truck, then perhaps the best way to accomplish this is to trick the upper layers into detecting many, many wheel shaped objects - in fact many more than would be expected even in true truck images. In this way there should be quantitative differences in the representations produced at intermediary layers - even though both ultimately end up being classified as the same class.

\section{Discussion\label{sec:Discussion}}
This paper has described a method for confusing state-of-the-art deep convolutional neural networks, by perturbing images from the ImageNet dataset in such a way that they can be reassigned to any other class leaving the images visually indistinguisible. It has also shown that creating a reassignment to a specific target label involves multiple gradient evaluations; i.e there is not a simple linear relationship to the nearby perturbation of an image that results specific labels being acquired. 

The existence of adversarial exemplars has strong implications for what exactly the networks we have been training are actually learning from the data. The fact that such simple methods described here can fool these networks suggests that the information contained about image classes on even large datasets like ImageNet are insufficient to allow for comprehensive, continuous generalisation. Simple tranformations (flipping, cropping, etc) can provide some resistance to pre-generated adversarial images, but this is far from a solution to the underlying problem. The current optimisation paradigms, network architectures or indeed both are fundamentally insufficient to resist these kinds of local instabilities. 

\bibliographystyle{plainnat}
\bibliography{bibtex}

\end{document}